\documentclass[sigconf,authorversion,nonacm]{acmart}
\usepackage{subcaption}
\usepackage{array}
\newcolumntype{?}{!{\vrule width 1pt}}

\AtBeginDocument{%
  }

\setcopyright{acmlicensed}
\copyrightyear{2026}
\acmYear{2026}
\setcopyright{cc}
\setcctype{by}
\acmConference[MMSports '26]{9th Int. Workshop on Multimedia Content Analysis in Sports}{November 10--14, 2026}{Rio de Janeiro, Brazil}
\acmBooktitle{9th Int. Workshop on Multimedia Content Analysis in Sports (MMSports '26), November 10--14, 2026, Rio de Janeiro, Brazil}
\acmDOI{10.1145/3841455.3841534}
\acmISBN{979-8-4007-2944-7/2026/11}

\begin{document}

\title{Event Detection in Table Tennis Videos using 2D Keypoints}

\author{Rainer Lienhart}
\authornote{All three authors contributed equally to this research.}
\authornote{A significant portion of the overall work was carried out during the research stay at the National Institute of Informatics, Toyko, Japan.}
\orcid{0000-0003-4007-6889}
\affiliation{%
	\institution{University of Augsburg}
	\city{Augsburg}
	\country{Germany}
}
\email{Rainer.Lienhart@uni-a.de}

\author{Daniel Kienzle}
\authornotemark[1] 
\orcid{0000-0001-7829-1256}
\affiliation{%
	\institution{University of Augsburg}
	\city{Augsburg}
	\country{Germany}
}
\email{Daniel.Kienzle@uni-a.de}

\author{Shin'ichi Satoh}
\orcid{0000-0001-6995-6447}
\affiliation{%
 	\institution{National Institute of Informatics}
 	\institution{University of Tokyo}
	\city{Tokyo}
 	\country{Japan}
}
\email{satoh@nii.ac.jp}

\author{Anastasiia Bilinska}
\authornotemark[1] 
\orcid{0009-0003-1326-7548}
\affiliation{%
	\institution{University of Augsburg}
	\city{Augsburg}
	\country{Germany}
}

\begin{abstract}
 This paper addresses the challenge of automatic, frame-accurate event detection in table tennis videos. Current methods for estimating 3d ball trajectories and ball spin typically require that key events, such as ball-racket contacts, have already been identified in advance. This requirement makes it difficult to apply these methods to longer, unedited video recordings. To overcome this limitation, we propose EventNet, a two-stage pipeline to detect key events:
 (1) 2d keypoints are extracted of the upper-body poses for both players, table corners and ball center. A small keypoint transformer combines them into a compact representation that is robust to changes in viewpoint, lighting, and background clutter. (2) The temporal sequences of these frame-based representations are processed by a transformer encoder that predicts two time-to-event values for each frame, indicating how close the current frame is to the next and previous ball-racket contact. One novelty is a new, temporal cosine-like target signal. Furthermore, we introduce viewpoint augmentation via 3D reprojection and frame-rate augmentation  to improve robustness and generalization. Our extensive ablation study gives deeper insights into the importance of various architectural and training aspects. Experimental results show that the proposed approach achieves an F1 score of 91.16\% and a mean frame deviation between ground truth and predicted frame  of 0.42 
 on the Latte-MV dataset and 73.08\% / 1.16 on the challenging TTHQ dataset.
 Overall, our work demonstrates that 2d keypoint-based temporal modeling with 
 our EventNet architecture is a promising and practical approach for automatic event detection in table tennis videos.
\end{abstract}

\ccsdesc[500]{Computing methodologies~Activity recognition and understanding}
\ccsdesc[500]{Computing methodologies~Visual content-based indexing and retrieval}

\keywords{frame-accurate event detection, ball-racket event detection, event detection in table tennis, Transformer-based event detection}

\begin{teaserfigure}
  \centering
 \includegraphics[width=0.9\textwidth]{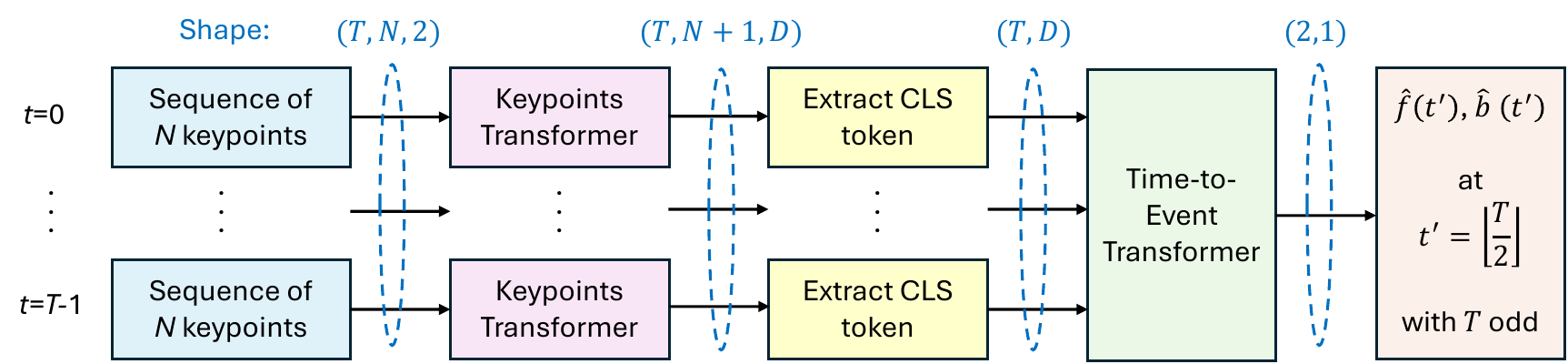}
  \caption{Overview of our EventNet: All keypoints per frame are individually condensed into a CLS token by the \emph{Keypoints Transformer}, before the \emph{Time-to-Event Transformer} determines the temporal distance to the previous and next event  ($b(f'), f(t')$).}
  \Description{Enjoying the baseball game from the third-base
  seats. Ichiro Suzuki preparing to bat.}
  \label{fig:overall_a}
\end{teaserfigure}

\maketitle

\section{Introduction}
In table tennis, recent works \cite{kienzle_uplifting_2026, KienzleSchoenLienhartetal.2025, etaat2025lattemvlearninganticipatetable} on 3d reconstruction and ball spin estimation have made it possible to analyze ball trajectories with high precision. However, these approaches usually assume that rallies are already segmented and timestamps of important events (such as ball-racket contacts or bounces on the table) are known in advance. In practice, this assumption is rarely satisfied, especially for long, uncut broadcast videos. 
As a result, existing 3D trajectory and spin estimation pipelines cannot yet be applied fully automatically to such unedited broadcast videos.

To address this limitation, this work focuses on automatic frame-accurate event detection such as ball-racket contact in table tennis videos using 2d keypoint representations instead of raw image data. We significantly extend and improve the work of \cite{einfalt2019frame} on frame-accurate event detection of foot ground contact in long jump and triple jump using 2d keypoint sequences.

Our proposed approach follows a two-step pipeline: First, 2d keypoints are extracted for the ball, the table, and the players' pose. The motion states of both players are represented through their upper body pose keypoints, the ball's motion through its location, and  the table keypoints provide geometric context. These keypoints in each frame are normalized and transformed by a keypoints transformer into a compact temporal sequence. Second, a time-to-event transformer then models the temporal dependencies in these sequences and predicts two time-to-event values for each frame, namely, how close a frame is to the previous and next event, which allows the model to locate ball-racket contact events in time precisely.

Several extended and novel design choices beyond our baseline model \cite{einfalt2019frame} are introduced to improve robustness, prediction performance, and generalization. Our key contributions are:

\begin{itemize}
	\item The proposed fps augmentation makes our models insensitive to varying, unknown frame rates of input videos.
	\item The proposed viewpoint augmentation through 3d reprojection mitigates the effect of the limited diversity of camera angles in the training dataset and simulates the complexity of real broadcast conditions. \\
	Both augmentations are possible due to our smart input representation as 2d keypoints without pixel information.
	\item We propose a relative linear and cosine-like time-to-event encoding, instead of the absolute linear event encoding of \cite{einfalt2019frame}. Both provide a smoother, more informative, and fps insensitive training signal. 
	\item The TCN backbone of \cite{einfalt2019frame} is replaced with a transformer-based architecture. This improves event detection performance at reduced training time by a factor of 0.32 to 0.48.
	\item An extensive ablation study gives deeper insights into the importance of various architectural and training aspects.		
\end{itemize}

\section{Related Work}
Previous work on event detection in sports videos can be roughly divided into three groups: methods that use (a) motion capture data, (b) work directly on raw video, or (c) rely on human poses and other keypoints as an intermediate representation.

Some early systems assume access to accurate 3D motion capture data \cite{vicente2016high,fernandez2013machine}. With special sensor- or marker-based setups, they detect key poses and split movements into actions, for example in martial arts or fitness exercises. These methods work well in controlled lab environments but need expensive hardware and are not suitable for large amounts of broadcast-like sports videos.

If only standard monocular videos are available, many approaches work directly on image or motion features \cite{li2010automatic,wu2002events}. Classical methods use silhouettes and motion-segmentation features for sports such as high diving or long jump, sometimes together with simple kinematic models. More recent deep learning methods train CNNs end-to-end on raw video frames to recognize actions or detect sparse events, for example in soccer or swimming \cite{giancola2018soccernet,sha2014understanding, victor2017continuous,woinoski2020towards}. These models do not need explicit motion representations, but they require a lot of annotated data and often generalize poorly when camera viewpoints or recording conditions change.

A different line of work uses 2d human pose sequences as a compact description of motion \cite{lienhart_mining_2018, yagi_estimation_2018, einfalt2019frame}. It has been shown that even noisy 2d keypoints can be used to estimate step frequency in sprinting, split athletic movements into phases, and detect sparse events in athletics videos with convolutional sequence models. \cite{einfalt2020decoupling} proposes to first estimate 2d poses with Mask R‑CNN and then apply pose-based sequence models for event detection in swimming and athletics. Their results show that pose sequences allow accurate and data-efficient event detection and can be used in practical systems for different sports.

This work follows the same general idea of separating visual perception from temporal reasoning, but focuses on table tennis. Compared to athletics and swimming, table tennis is more challenging because the ball is small, often occluded, filmed from many different viewpoints and moves fast.

\section{Approach}
Our goal is to perform frame-accurate event detection in video recordings of table tennis matches. As a representative event, we choose the moment of ball-racket contact, i.e., we want to predict all frames showing the desired event in long, unedited broadcast videos of table tennis matches. We assume that valid results of a 2d ball, table tennis corner and net detector, as well as a human pose estimator are available, providing us most of the time throughout a video with an estimate of the 2d pixel positions of the ball center, the table and net corners, as well as the keypoints of the torsos and upper limbs of both table tennis players\footnote{The source code and models are available at \url{ https://github.com/MLCVLab/EventNet}}. 

\subsection{Baseline Model}
The work in \cite{einfalt2019frame} serves as our baseline model. It addresses frame-accurate event detection in long and triple jump videos.
This method relies purely on 2d pose sequences extracted from the athlete in each frame to determine the beginning or end of foot ground contact. Pose sequences are a very compact and better suited representation for learning motion patterns over time than raw videos. 

Adapted to our application scenario, 
each frame is described by a concatenated vector of 2d keypoints consisting of the pixel positions of the ball center, the table and net corners as well as the keypoints of the torsos and upper limbs of both table tennis players. Keypoints without valid values for whatever reason (e.g., not in the field-of-view or the detector missed it) are encoded as NaN values. In \cite{einfalt2019frame} the sequence of keypoints is fed into a fully \underline{T}emporal \underline{C}onvolutional \underline{N}etwork (TCN \cite{bai_empirical_2018}) that predicts the frame distance to the next and previous occurrence of the desired event type. 

A key contribution of \cite{einfalt2019frame} is its event detection formulation. Instead of classifying each frame directly as event or non-event, it defines for every frame two target values for each event type: a forward and a backward time-to-event indicator value. The forward/backward value tells the network how far the current frame is away from the next/previous event. By predicting continuously time-to-event values (i.e., for every frame) instead of classifying a few rare events, a severe class imbalance at training time is avoided. 

The TCN model \cite{bai_empirical_2018}
is based on temporal convolutional blocks. Each block uses dilated 1D convolutions over time, followed by batch normalization, ReLU, dropout, and residual connections. The network takes flattened keypoint locations as input, and outputs the forward and backward timing indicators for all event types at each time step. During training, the input length is set to the model’s receptive field, and the prediction is made for the center frame of the sequence. At inference time, the network processes the full pose sequence of a video and predicts time-to-event indicator values for every frame. Event locations are detected at time steps $t$ where both indicator values simultaneously identify the same frame as the next and previous event.

\subsection{Input: Sequence of 2d Keypoint Sequences}
As already said, we assume that we have for every frame $i$ (estimates of) the 2d pixel positions of the ball center, the table and net corners, as well as the keypoints of the torsos and upper limbs of both table tennis players. Missing keypoint values (either not visible in the frame or missed by the detectors) are encoded by NaN values.
For each frame $i$, we construct its feature tensor $\mathbf{x}_i$ of shape $N \times 2$ by concatenating the ball location with the six table keypoints (four table corners and two net points) and a selected subset of $b$ body keypoints of each player, resulting in $N = 1 + 6 + 2b$ keypoints per frame (see Fig. \ref{fig:keypoints_sequence}). We construct  
two 
different, named keypoint sets each with a different subset $b$ of upper body keypoints: 
\begin{enumerate}
	\item UB = \emph{upper body} = \{ball, table, head, shoulders, elbows, wrists, hips\}, $N=25$
	\item NO = \emph{no body points} = \{ball, table\}, $N=7$
\end{enumerate}
We will evaluate experimentally in Section \ref{expResult} which keypoint set is more informative for the task at hand. 

\paragraph{Raw and Normalized Input Sequence } All raw\footnote{The attribute \textit{raw} indicates that all 2d coordinates are given in the unnormalized, original pixel space of each video frame.} input sequences $\mathbf{x}$
of length $T$
are of shape $T \times N \times 2$. As the training will be conducted using the LATTE-MV dataset \cite{etaat2025lattemvlearninganticipatetable}, all player keypoints as well as the table and ball coordinates are annotated for each frame (including NaN values for not available keypoints). 
However, the raw 2d keypoint coordinates 
vary widely in scale across different videos due to differences in camera distance to the tennis table, viewpoint, scene geometry, and video resolution. To make the  representation of $\mathbf{x}$ invariant to these factors, all keypoint coordinates in each input sequence $\mathbf{x}$ are normalized to the range of $[-1, 1]$ as follows: 
\begin{itemize}
	\item Missing or failed detections (NaN values) in $\mathbf{x}$ are  replaced by 0.0 in max or by $width_{video}$ in min operations. 
	\item  We compute the global minimum $m = \min(\mathbf{x})$  and maximum  $ M = \max(\mathbf{x})$ over each complete $T \times N \times 2$ tensor.
	\item  Each keypoint is then scaled and shifted into the interval $[-1, 1]$ via 
	$\mathbf{x}''= 2 \cdot \mathbf{x}'- 1 \text{  with  } \mathbf{x}' = \frac{\mathbf{x} - m}{M - m}$.

\end{itemize}

This sequence-wise normalization ensures that all coordinates lie in a consistent bounded range.

\subsection{Target: Time-to-Event Indicator Signal}
Each training and test video is fully annotated with a temporally ordered set $\mathbf{e}=\{ e_1, \dots, e_E \}$ of size $E$ of event occurrences of type $c$, with $c=\text{"ball-racket contact"}$. 
Each $e_i \in \{1, \dots, T\}$  specifies the frame index of a ball-racket contact event by any player. $T$ is the temporal length of the input video in frames. 
Our task is now to construct two useful time-to-event indicator signals named $f_c(t)$ 
and $b_c(t)$, that encode for each frame the 
distance to the next  and from the previous event of type $c$. This concept 
is taken from~\cite{einfalt2019frame}. 

\begin{figure}[htbp]
	\centering
	
	\begin{subfigure}[b]{0.48\textwidth}
		\centering
		\includegraphics[width=\textwidth]{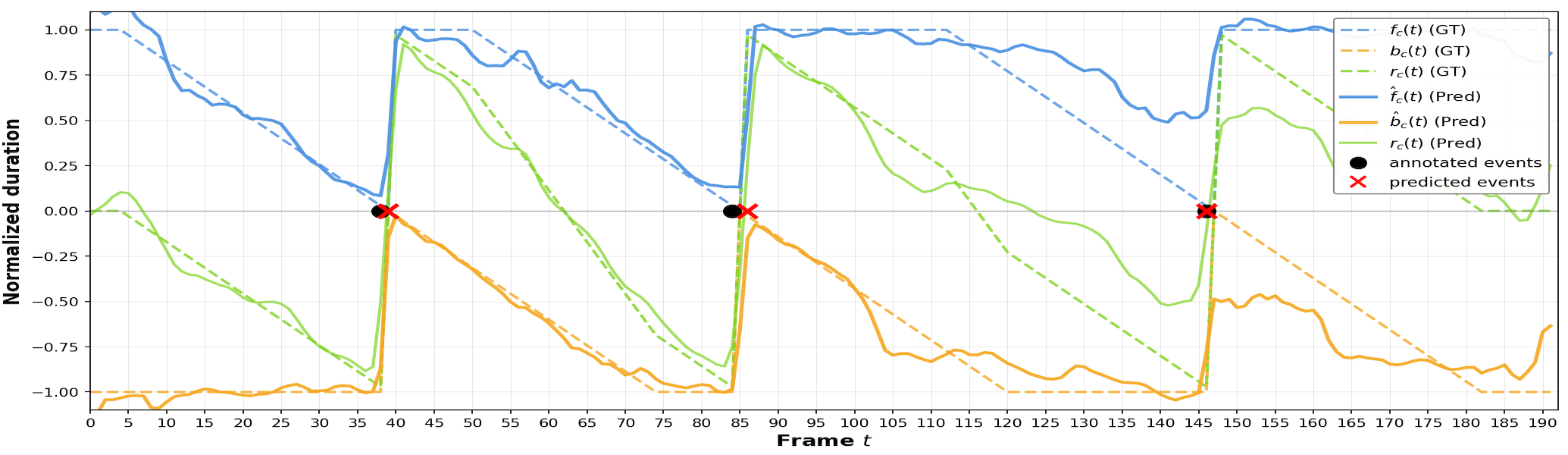}
		\caption{Absolute linear time-to-event encoding (short "\emph{absLin}")}
		\label{fig:abs_linear}
	\end{subfigure}
	
	
	\begin{subfigure}[b]{0.48\textwidth}
		\centering
		\includegraphics[width=\textwidth]{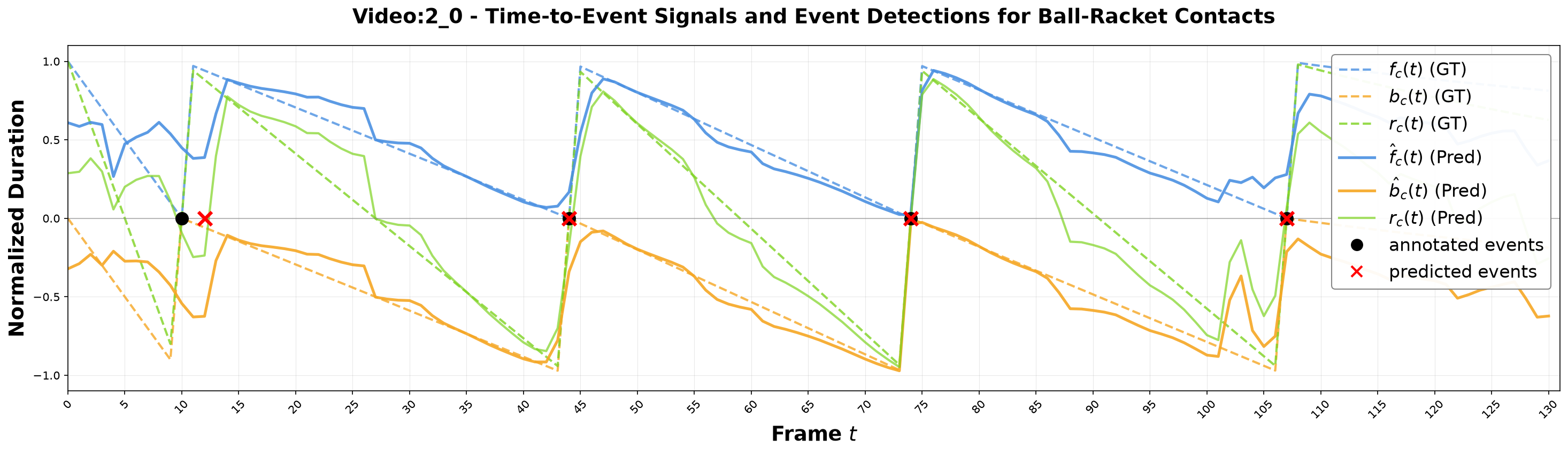}
		\caption{Relative linear time-to-event encoding (short "\emph{relLin}")}
		\label{fig:rel_linear}
	\end{subfigure}
	
	\begin{subfigure}[b]{0.48\textwidth}
		\centering
		\includegraphics[width=\textwidth]{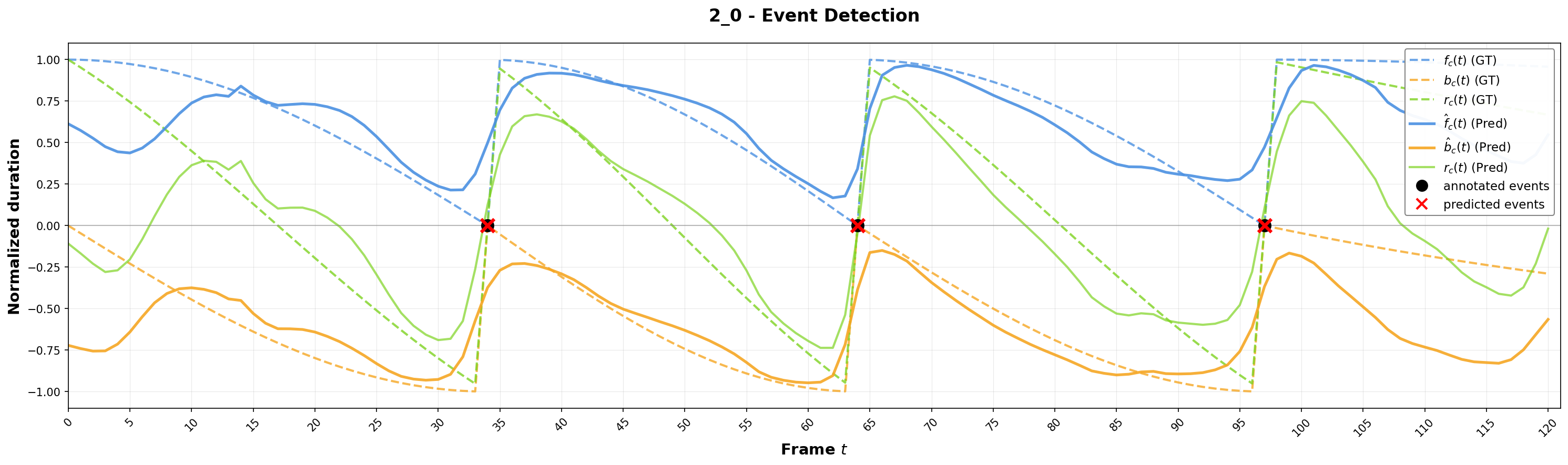}
		\caption{Cosine-like time-to-event encoding (short "\emph{cos}")}
		\label{fig:use_sin}
	\end{subfigure}
	
	\caption{Visualizations of our three variants of time-to-event target signals: (a) absolute linear with $t_{max} = 35$, (b) relative linear,  and (c) cosine-like. Dashed lines show the target time-to-event signal: forward signal $f_c(t)$ in orange, backward signal $b_c(t)$ in blue, summed signal $r_c(t)$ in green. Solid lines show the predictions of one of our trained models.}
	\label{fig:target_signal}
\end{figure}

\paragraph{Absolute Linear Time-to-Event Signal}
In~\cite{einfalt2019frame} the target temporal distances to the events are saturated normalized by a constant $t_{\max}$ to ensure that the target values lie within the range $[0,1]$ for $f_c(t)$ and $[-1,0]$ for $b_c(t)$. 
In practice, parameter $t_{\max}$ proved difficult to tune in our task due to different fps and highly variable ball speeds. If $t_{\max}$ is too 
small, only frames near events show meaningful values between $0$ 
or $1$, while distant frames at time $t$ saturate to the extreme value of  
$f_c(t)=1$ and $b_c(t)=-1$, respectively. If $t_{\max}$ is 
chosen too large, the signal becomes nearly flat. These issues made learning challenging as clipped signals give no feedback during learning. The dashed lines in Fig.~\ref{fig:abs_linear} show a target signal with flat regions.

\paragraph{ Relative Linear and Cosine-Like Time-to-Event Encoding}
To address this shortcoming, we propose two new encodings. Both make sure that every event, regardless of the sequence's fps or the ball's speed, is represented by the exact same mathematical pattern between adjacent events. Instead of measuring the absolute distance to an event,
the signal provides a relative temporal position between adjacent events. 
For each time index $t$, the times of the closest previous and 
next event are defined as:
$	t_{\mathrm{prev}}(t) = 
	\max \{ e_i \in \mathbf{e} \mid e_i \le t \}, \\
	t_{\mathrm{next}}(t) = 
	\min \{ e_i \in \mathbf{e} \mid e_i \ge t \}$.
To ensure that the formulation remains well-defined at the boundaries of a video sequence, we apply the following rules: if no previous event exists (i.e., the current frame lies before the first annotated event), we set $t_{\mathrm{prev}}(t) = 0$. Similarly, if no 
future event exists (i.e., the current frame lies after the last 
annotated event), $t_{\mathrm{next}}(t) = t + \Delta$, where 
$\Delta > 0$ is a fixed constant. This avoids undefined 
operations when computing distances to neighboring events and ensures 
stable and bounded  timing indicators at the beginning 
and end of a sequence. 

Based on these neighboring events, we define relative linear (relLin) and cosine-like (cos) time-to-event encodings:
\begin{equation}
	f_{\text{relLin}}(t) = 
	\frac{t_{\mathrm{next}}(t) - t}{t_{\text{next}} - t_{\text{prev}}}, 
	b_{\text{relLin}}(t) = 
	-\frac{t - t_{\mathrm{prev}}(t)}{t_{\text{next}} - t_{\text{prev}}} \\
\end{equation}
\begin{equation}
	f_{\text{cos}}(t) = 
	\sin\!\left(
	\frac{\pi}{2} 
	\frac{t_{\mathrm{next}}(t) - t}{t_{\text{next}} - t_{\text{prev}}}
	\right), 
	b_{\text{cos}}(t) = 
	-\sin\!\left(
	\frac{\pi}{2} 
	\frac{t - t_{\mathrm{prev}}(t)}{t_{\text{next}} - t_{\text{prev}}}
	\right)
\end{equation}
By construction both encodings satisfy
$f(e_i)\hspace{-3pt}=\hspace{-2pt}b(e_i)\hspace{-3pt}=\hspace{-2pt}0$ $\forall e_i\in \mathbf{e}$, 
while they take smooth, non-zero values between adjacent events. The pair $(f(t), b(t))$ thus forms a continuous encoding of the temporal distance to the next event, with $f(t)$ decreasing towards zero when approaching the next event and $b(t)$ increasing towards zero after the previous event. 
Unlike the relative linear encoding (dashed lines in Fig.~\ref{fig:rel_linear}), which does not consider that the prediction task gets more difficult and uncertain the further the next/previous event is away, the cosine-like signal (dashed lines in Fig.~\ref{fig:use_sin}) somehow captures that by defining a more similar value the further it is away from the next/previous event. We will evaluate experimentally in Section \ref{expResult} which temporal time-to-event encoding works best.

\subsection{EventNet Architecture}

\begin{figure}[htbp]
	\centering
	\includegraphics[width=0.48\textwidth]{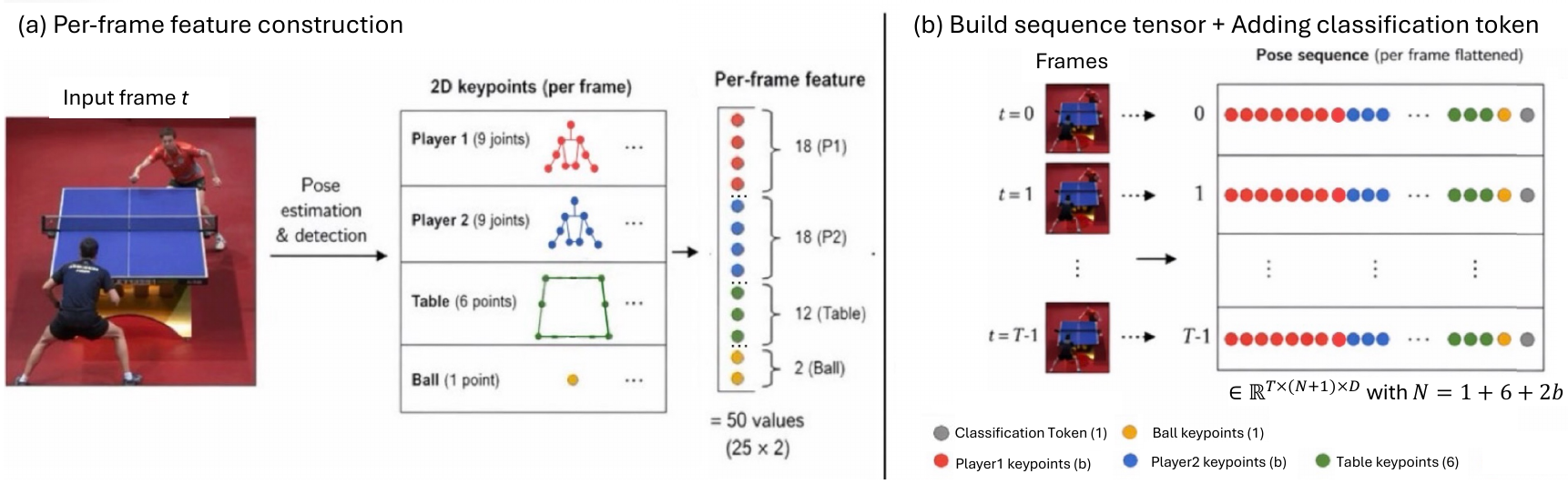}
	\caption{Construction of the input representation. 
		(a) Keypoints from both players, the table, and the ball are extracted and concatenated. Here, 9 per player, 6 for the table, and 1 for the ball, resulting in 25 $(x, y)$ keypoint coordinates per frame. 
		(b) EventNet input by assembling the pose sequence across frames (projection each keypoint to size $D$) and adding the CLS token of size $D$ to each frame.}
	\label{fig:keypoints_sequence}
\end{figure}

In our EventNet, we replace the TCN from the baseline model \cite{einfalt2019frame} with an encoder-only transformer architecture. It consists of a \emph{Keypoints Transformer} which converts the sequence of 2d keypoints for each frame into a compact per-frame embedding, and a \emph{Time-to-Event Transformer} which models the temporal evolution of these embeddings to predict the values of the time-to-event indicators $f(t)$ and $b(t)$ for time instances $t$.
Fig.~\ref{fig:overall_a} illustrates this architecture. 
The input to EventNet is a tensor $\mathbf{x} \in \mathbb{R}^{T \times N \times 2}$, where $T$ denotes the number of frames in the sequence, and $N$ the number of 2d keypoints per frame (see Fig. \ref{fig:keypoints_sequence}), and the last dimension provides the normalized $(x,y)$ coordinates of each keypoint. 

\begin{figure}[t]
	\centering
	\includegraphics[width=0.9\linewidth]{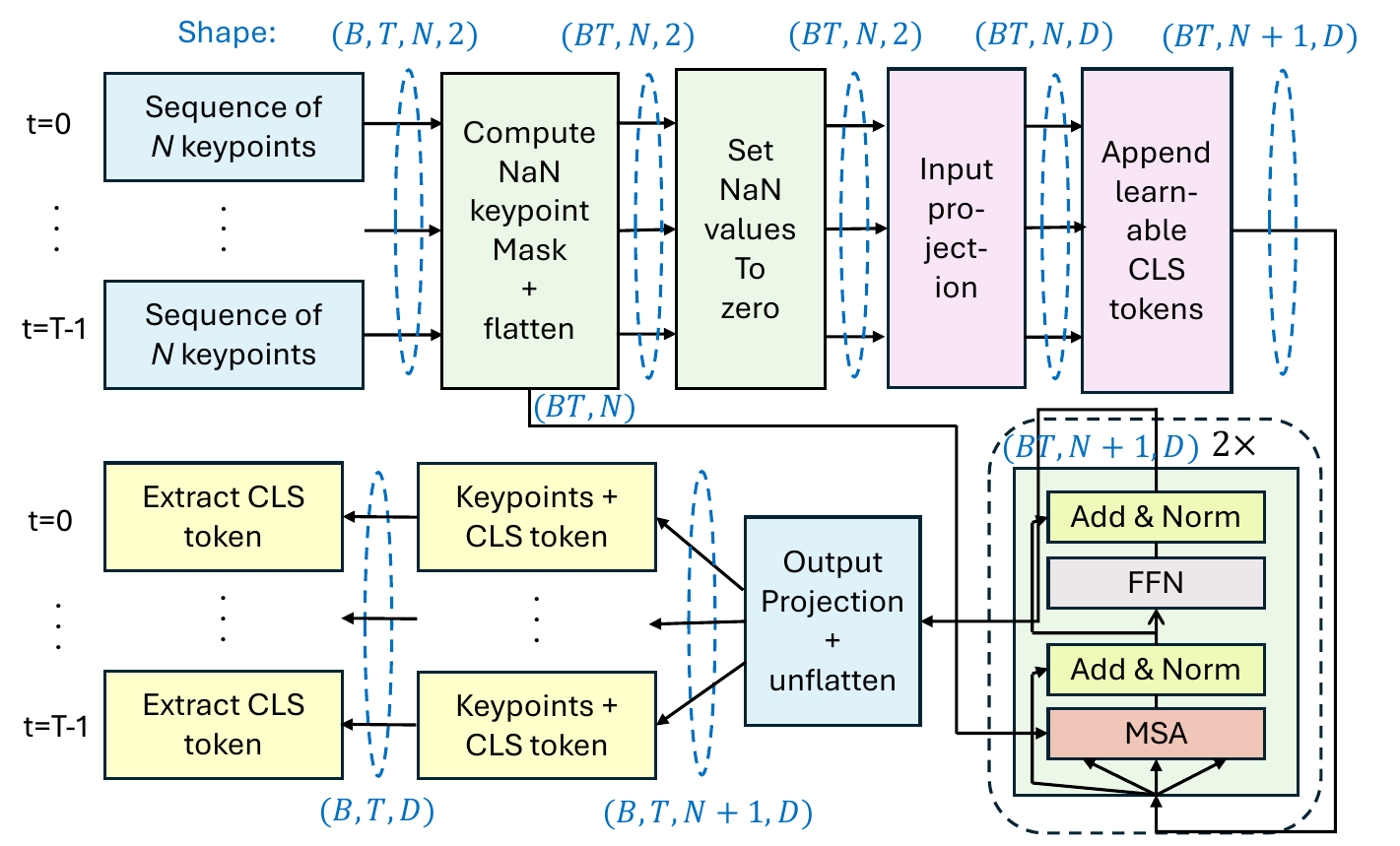}
	\caption{Overview of the Keypoints Transformer architecture. It shows how a sequence of keypoints per frame is transformed into a CLS token and prepared as input for the temporal Time-to-Event Transformer.}
	\label{fig:keypoints_transformer}
\end{figure}

\paragraph{Keypoints Transformer}Since some frames may have missing keypoints, represented as NaN values, these coordinates are handled explicitly
by the Keypoints Transformer as illustrated in Fig.~\ref{fig:keypoints_transformer} top. 
For each frame, a binary mask marks all keypoints with NaN coordinates as missing. 
Then, each of the $N$ keypoints is projected linearly to a latent dimension $D$. The result is masked again so that no missing keypoint can contribute to the representation. To obtain a fixed-size representation of the keypoints for each frame, a learnable $\text{CLS}$ token is appended to the keypoint sequence. The resulting sequence of $N+1$ keypoints is processed by a small stack of two transformer encoder blocks with multihead self-attention. Again, the attention is masked to ensure that NaN keypoints do not influence the self-attention at all. After the encoder, only the CLS output token is retained for each frame as the compact descriptor of the entire keypoint set at time $t$. This results in an output sequence $\mathbf{z} = (\mathbf{z} _1, \dots, \mathbf{z} _T) \text{ with } \mathbf{z} _t \in \mathbb{R}^D$  of shape $(T, D)$, which summarizes the spatial relationships among all visible keypoints over time, while remaining robust to missing values. 

Alternatively, we also implement one variant of Keypoints Transformer where a mask token for each keypoint type is learned and used if an associated keypoint contains NaN values. We will evaluate experimentally in Sec. \ref{expResult} which approach of handling NaN keypoints is superior.

\begin{figure}[t]
	\centering
	\includegraphics[width=1\linewidth]{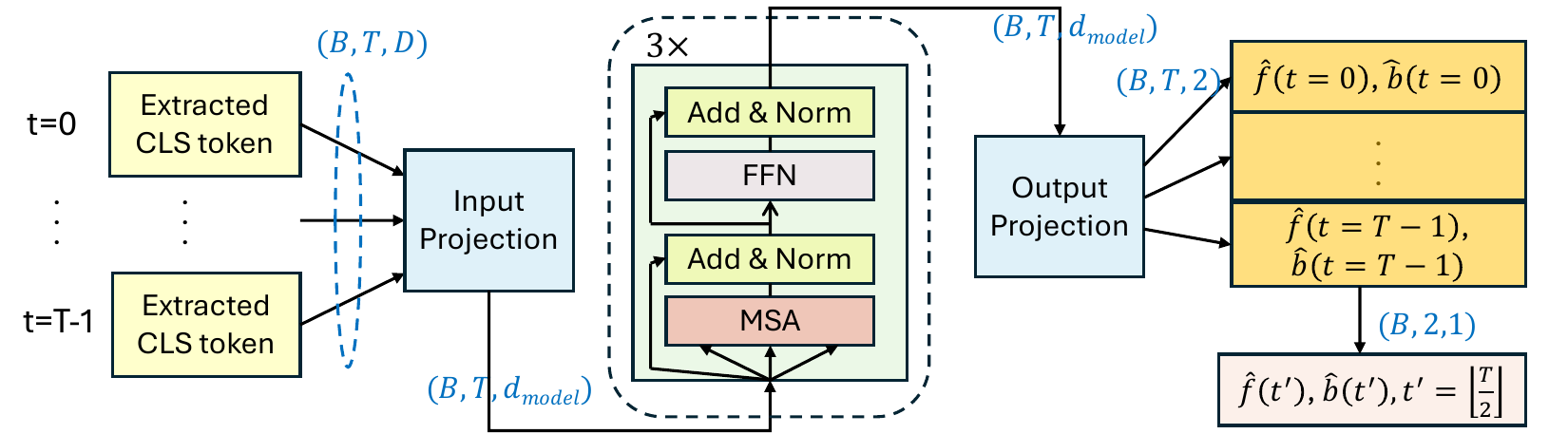}
	\caption{Architecture of the Time-to-Event Transformer that takes a CLS token sequence and produces time-to-event indicators $f(t)$ and $b(t)$.
	}
	\label{fig:main_transformer}
\end{figure}

\paragraph{Time-to-Event Transformer} The summarized representation of 2d keypoints, obtained as CLS token sequence $\mathbf{z}$,
is then passed to the Time-to-Event Transformer, shown in Fig.~\ref{fig:main_transformer}. Here, a stack of three transformer encoder blocks models dependencies across time using multi-head self-attention, residual connections, pre-norm layer normalization, and a position-wise feed-forward network with ReLU activation\footnote{In the experimental results section, we will also explore the effect of replacing ReLU by GELU or the whole FFN-block by SwiGLUFFN.}. 
In this stage, the model captures long-range temporal context across the sequence of per-frame CLS tokens and maps them frame-wise to forward and backward time-to-event indicators. A final linear layer projects each temporal embedding to the 2d output values corresponding to the predicted $f(t)$ and $b(t)$ signals. As the prediction targets depend only on the nearest neighboring events, the model is trained on fixed-length sequences and predicts timing indicators for the central time step $m=\lfloor T/2 \rfloor$. The model is optimized with Huber loss applied to the predicted forward and backward indicators. 

\subsection{Event Extraction}
With perfect predictions, all events could be identified at time steps $t$ where 
$f(t)=b(t)=0$. In practice, however, it is challenging for a model to perfectly reproduce the target signal, and thus $f(t)$ and $b(t)$ rarely reach zero. Therefore, both indicators are combined into a single discriminative event indicator $r(t)$ defined as
\begin{equation}
r(t) = f(t) + b(t). 
\end{equation}
Events are then localized in $r(t)$ in three steps:

\begin{enumerate}
	\item \textbf{Zero-crossings with a sufficiently positive slope:} Candidate events are found 
	at zero-crossings where $r(t)$ surpasses upward a threshold of $\Delta h$. The threshold acts as a confidence gate: true events produce strong zero-crossings in $r(t)$ that reliably exceed $\Delta h$, while noise-related predictions remain below it and are filtered out.
	
	\item \textbf{Linear interpolation:} 
	The zero-crossing location of each candidate event at time $t$ is refined by linear interpolation. 
	For consecutive frames where $r(t)<0$ and $r(t+1)>0$, the 
	exact zero-crossing $t^*$ is computed as
	\begin{equation}
	t^* = t- r(t) \cdot \frac{(t+1) - t}{r(t+1) - r(t)} = t_i - \frac{r(t) }{r(t+1) - r(t)},
	\end{equation}
	then rounded to the nearest frame index, yielding a refined  
	time of the event candidate with associated confidence scores using $s=r(t+1)-r(t)$.
	
	
	\item \textbf{Non-maximum suppression (NMS):} We apply NMS to eliminate redundant event candidates by sorting all event candidates according to their score $s$ and then -- starting from the highest scoring -- iteratively eliminating all event candidates that are within a temporal window of $\Delta t$. This eliminates redundant detections of the same event.
\end{enumerate}

Fig.~\ref{fig:use_sin} illustrates an example of the network output together with the combined event indicator $r(t)$. Event occurrences are identified at $r_c(t) \approx 0$, where the signal changes sign from negative to positive. The first peak occurring around the $10^{th}$ frame does not reach the predefined threshold of $\Delta h=0.5$, thereby preventing a false positive prediction.

\subsection{Viewpoint Augmentation} 
The LATTE-MV \cite{etaat2025lattemvlearninganticipatetable} dataset is limited to videos captured from a single, side-view camera viewpoint. This results in a strong viewpoint bias, as real-world table tennis footage exhibits various camera angles and focal lengths.
To improve generalization to unseen viewpoints, the training data is augmented by generating synthetic cameras using the available 3D reconstructions. For each sequence, a realistic camera is created as follows: The intrinsic parameters are defined by randomly sampling focal lengths 
and the principal point within realistic ranges to construct the intrinsic matrix $M_{\mathrm{int}}$. The extrinsic parameters are obtained by sampling the camera position in spherical coordinates, including the distance from the table as well as azimuth and elevation angles, together with a look-at 
point near the table center. From these parameters, the camera position 
$\mathbf{c}$, forward direction $\mathbf{f}$, right vector $\mathbf{r}$, and 
up vector $\mathbf{u}$ are computed. The orientation is constrained such 
that the up vector has a positive $z$-component, ensuring a physically 
plausible configuration. These components are then used to construct the 
extrinsic matrix $M_{\mathrm{ext}}$.

Each generated camera is validated by projecting the full 3D sequence. 
A camera pose and position is accepted only if more than $95\%$ of all keypoints remain within 
the image bounds and the projected scene spans a sufficient portion of 
the image area. Cameras that do not satisfy these criteria are discarded 
and resampled.
Finally, all valid 3D keypoints (players, table, and ball) are projected 
onto the synthetic camera’s 2d image plane using the standard projection 
equations
\begin{align}
\mathbf{r}_{\mathrm{cam}} 
&= \mathrm{world2cam}(\mathbf{r}_{\mathrm{world}}, M_{\mathrm{ext}}) \\
\mathbf{r}_{\mathrm{img}} 
&= 
\mathrm{cam2img}(\mathbf{r}_{\mathrm{cam}}, M_{\mathrm{int}}).
\end{align}
These sequences with event annotations are used in training.

\subsection{Fps Augmentation}
In the Latte-MV dataset, all videos are recorded at 30 fps. While this is convenient, in real-world scenarios, however, table tennis videos are captured at various frame rates. Thus, the training data was augmented with sequences at diverse fps.

For every pose, ball, and table keypoint, the same interpolation scheme is applied. Given $N_{in}$ frames at an input frame rate $f_{\text{in}}$ sampled at time instances $t^{\text{in}}_i = \frac{i}{f_{\text{in}}}, \text{ } i = 0, 1, \ldots, N_{in}-1$, a new time grid at output frame rate $f_{\text{out}}$ is constructed at time instances $t^{\text{out}}_j = \frac{j}{f_{\text{out}}}, \text{} j = 0, 1, \ldots, N_{out}-1$ by preserving the total duration T:
\begin{equation}
	T = t^{\text{in}}_{N-1} = \frac{N_{in}-1}{f_{\text{in}}} =  \frac{N_{out}-1}{f_{\text{out}}}.
\end{equation}
Thus, 
	$N_{\text{out}} = \left\lfloor T \cdot f_{\text{out}} \right\rfloor + 1$.
Each input dimension $d$ of the keypoints is then interpolated independently using linear interpolation.  

The target values for the new sampling rate must be derived from the event array $event_c^{\text{in}} \in \{0,1\}^{N_{in} \times 2}$ which encodes two types of events: ball-racket contact of player 1 and player 2. However, linear interpolation is not possible with discrete events. It would produce physically meaningless values. 
Instead, for each column $c$, the set $\mathcal{E}_c$ of event frame indices where a contact with the racket occurs is first extracted.
Each event index is then converted to an absolute timestamp in seconds: $\tau_e = \frac{e}{f_{\text{in}}} \text{ } \forall e \in \mathcal{E}_c$.
The timestamp is mapped to the nearest frame index in the new time grid:
\begin{equation}
	e^{\text{out}} = \mathrm{clip}\!\left(   \text{ }  \mathrm{round}\!\left( \tau_e \cdot f_{\text{out}} \right),\ 0,\ N_{\text{out}}-1 \right).
\end{equation}
The clipping ensures that the index remains within the valid range.
A new zero array $event_c^{\text{out}} \in \{0,1\}^{N_{\text{out}} \times 2}$ is then constructed, and each remapped event is placed at its new position:
\begin{equation}
	event_c^{\text{out}}[e^{\text{out}}, c] = 1.
\end{equation}

\section{Experimental Results}
\label{expResult}

\subsection{Datasets}

\paragraph{Latte-MV} All configurations of our approach were trained and evaluated on the Latte-MV dataset \cite{etaat2025lattemvlearninganticipatetable}. The dataset provides annotations of racket-ball contacts  as well as 2d and 3d keypoint annotations for ball, tennis table, and human poses of both players. It consists of 9,803 training, 2,099 validation, and 2,103 test videos. In case of viewpoint augmentation using 3D reprojection, the augmentation strategy was only applied to a subset of 2,328 training and 498 validation videos, for which the 3d ball coordinates were valid (i.e., not showing NaN values) at least 70\% of all frames. The remaining training and validation videos were used without augmentations. For fps augmentation, the frame rates varied between 30 and 60 fps. The test set was never unchanged. The keypoint annotations 
were rescaled to a resolution of $1920 \times 1080$  before training. Models were trained on a sequence length of $T = 29$, each configuration 3 times with 3 different seeds.

\begin{figure}[t]
	\centering
	\includegraphics[width=0.46\textwidth]{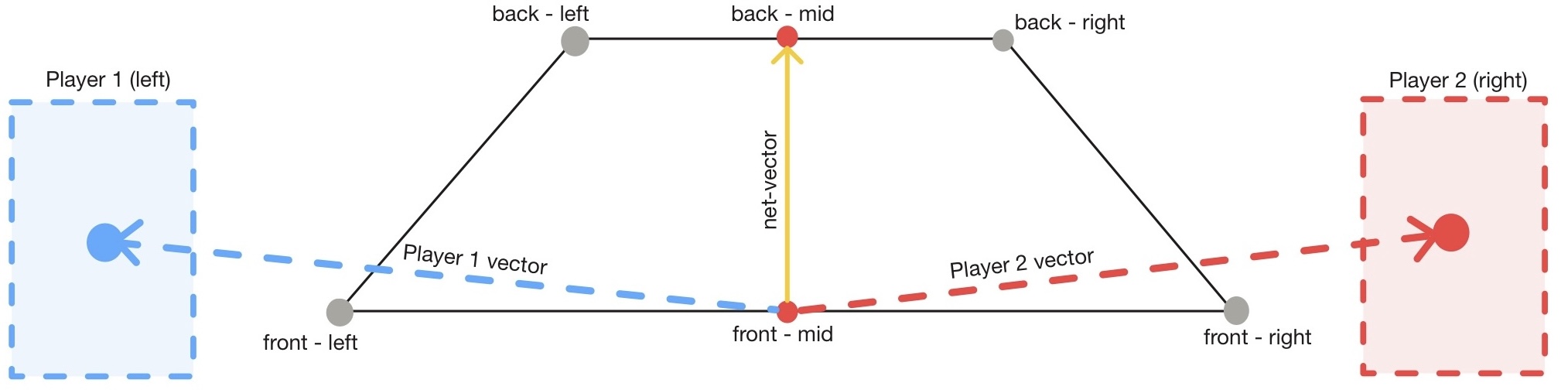}
	\caption{Spatial player assignment logic using table geometry. The net line is defined by the vector between the front-mid and back-mid keypoints. Player identities are assigned by calculating the 2d cross product between this net vector and the vector pointing to each player's bounding box center. The sign of the resulting scalar determines the "left" as Player 1 and "right" as Player 2, ensuring identity consistency despite camera perspective or player movement.}
	\label{fig:table_figure}
\end{figure}

\paragraph{TTHQ} The TTHQ dataset~\cite{kienzle_uplifting_2026} was used to assess the model on real-world table tennis footage recorded from very diverse viewpoints and a wide variety of scenes. All videos consist of standard or semi-broadcast YouTube content, captured at a resolution of $1920 \times 1080$ with a static camera setup per video. The frame rates range from 25 to 60 fps. From the TTHQ dataset, we only select videos containing two players, as our models are trained on two-player scenes. The final selection consists of six long videos, which were split into shorter clips and recorded under varying camera viewpoints and frame rates, resulting in a total of 56 clips.
TTHQ provides annotations of ball-racket contacts and tennis table corners, but not for the 2d poses of the two players and only some for ball positions. Therefore, the players were first localized in each video frame using RT-DETRv2 \cite{lv2024rtdetrv2improvedbaselinebagoffreebies} (model \texttt{rtdetr\_r50vd\_coco\_o365} \footnote{https://github.com/lyuwenyu/RT-DETR}). Then, since table tennis involves two players in our setup, our algorithm selects the two largest detected persons in each frame, followed by 2d pose estimation using a ViT-Pose-B model \cite{xu2022vitpose} (model \texttt{vitpose-base-simple} \footnote{https://github.com/ViTAE-Transformer/ViTPose}). 
To ensure consistent player assignment across a video, the two detected players are ordered in the input keypoints vector based on their spatial position relative to the table net (see Fig.~\ref{fig:table_figure}). First, the front-mid and back-mid table keypoints construct a line representing the table net. Next, the center point of each detected player is computed from his/her bounding box. The sign of the cross product between the net direction vector and the vector from the front-mid net point to each player determines whether a player is located on the “left” or “right” side of the table. 

From the full set of predicted pose keypoints, only a subset relevant to the upper body is used. These are:
\textit{head, left/right shoulder, left/right elbow, left/right wrist, and left/right hip}. 
Thus, each player is represented by nine 2d keypoints. 
The table and ball were detected in each frame using a SegFormer++ B0 model \cite{xie2021segformer, KienzleKantonisSchoenetal.2024}, a lightweight transformer-based segmentation model. It provides efficient and accurate per-frame segmentation, which makes it suitable for detecting small objects such as the ball and for localizing the table tennis table consistently across video frames.

\subsection{Architecture details}

The temporal transformer was trained with a default batch size of 200 for 30 epochs. The model parameters are optimized using the Adam optimizer. The initial learning rate is set to $5 \cdot 10^{-4}$, and a learning rate scheduler is used to improve convergence by reducing the learning rate by a factor of $\gamma = 0.3$ after epoch 2, allowing the models to fine-tune their weights with smaller updates in later training stages. Formally, the learning rate at epoch $t$ is given by $\text{lr}_t = \gamma^k \cdot \text{lr}$,
where $k$ is the number of milestones reached (in this case, $k=1$ after epoch 2). This scheduling strategy helps to stabilize training and improve overall performance. 

In addition to the current state of the time-to-event transformer, an exponential moving average (EMA) of its parameters is maintained during training, and all evaluations and inference experiments are performed using these EMA weights \cite{mean_teacher2017}.
The EMA model tracks a smoothed version of the weights, updating according to:
\begin{equation}
\theta_{\text{EMA}} \gets \alpha \, \theta_{\text{EMA}} + (1 - \alpha) \, \theta_{\text{current}},
\end{equation}
where $\theta_{\text{EMA}}$ are the EMA parameters, $\theta_{\text{current}}$ the current model parameters, and $\alpha = 0.999$ the decay factor. 
This approach stabilizes training by reducing the effect of noisy gradient updates and improves evaluation and inference performance. 

Two encoder layers with a classification token of dimension $d = 128$ were used to process keypoints. The time-to-event transformer uses three encoder layers, and the output consists of two predicted values per frame: $\hat{f}(t)$ (predicted forward timing indicator) and $\hat{b}(t)$ (predicted backward timing indicator).
The combined event indicator $\hat{r}(t)$ was used for discrete event extraction as described in Section~3.5, 
with non-maximum suppression applied using a window size of $w = 7$ frames and a minimal peak height of $0.5$.

\subsection{Evaluation metrics}
For evaluation, we let a trained model process a complete test video set to predict all ball-racket contact events. We assign to each ground truth event in the test set the closest predicted event if and only if it is within an absolute temporal distance of $\Delta t$ frames. Thus, a \textit{predicted event is considered correct} if (a) its absolute temporal distance to the assigned ground truth does not exceed the maximum allowed frame distance $\Delta t$ and (b) there is no other closer predicted event. In our experiments, we have chosen a default threshold of $\Delta t = 3$ frames.

Having defined what correct event detections are, we can now compute \textit{precision, recall}, and  \textit{F1} score (in percentage) over a complete test video dataset. The F1 score is the harmonic mean of precision and recall. Instead of a simple average, a harmonic mean punishes extreme failures. Thus, F1 will be our main performance score.
Additionally, the \textit{mean absolute frame deviation} (short \textit{L1 Diff}) of all correctly detected events (i.e., true positives) is also reported. A value of, for instance, 0.5 means that on average the error is 1/2 of a frame.

\begin{table*}
	\caption{Test results on Latte-MV and TTHQ using our EventNet. The models are exclusively trained 3 times with Adam on Latte-MV using fps and viewpoint augmentation, a sequence length of $T=29$ and a maximum allowed frame distance of $\Delta t = 3$. The temporal encoding and the non-linearity used in the FFN blocks of all transformer layers are set as specified for each configuration. "L1 Diff" stands for mean absolute frame distance between predicted and ground truth event frames.}
	  \label{tab:table_01}
  \begin{tabular}{l?cc|cc?cc|cc}
    \toprule
    & \multicolumn{4}{c?}{Mask all NaN Values} & \multicolumn{4}{c}{Mask Tokens for NaN Values} \\
    & \multicolumn{2}{c}{Latte-MV } & \multicolumn{2}{c?}{TTHQ} & \multicolumn{2}{c}{Latte-MV } & \multicolumn{2}{c}{TTHQ} \\
    Configuration & F1 Score & L1 Diff & F1 Score & L1 Dff & F1 Score & L1 Diff & F1 Score & L1 Diff \\
    \midrule
    NO, relative linear, ReLU & 89.51{ \small $\pm$0.67 } & 0.42{ \small $\pm$0.02 } & 69.25{ \small $\pm$2.75 } & 1.14{ \small $\pm$0.05 } & 89.58{ \small $\pm$1.00 } & 0.38{ \small $\pm$0.01 } & 73.08{ \small $\pm$3.92 } & 1.16{ \small $\pm$0.01 } \\ 
    NO, relative linear, GELU & 89.06{ \small $\pm$1.56 } & 0.42{ \small $\pm$0.00 } & 70.33{ \small $\pm$5.44 } & 1.13{ \small $\pm$0.05 } & 88.99{ \small $\pm$2.30 } & 0.41{ \small $\pm$0.04 } & 68.69{ \small $\pm$0.63 } & 1.18{ \small $\pm$0.10 } \\ 
    NO, relative linear, SwiGLUFFN & 88.44{ \small $\pm$1.32 } & 0.43{ \small $\pm$0.02 } & 68.72{ \small $\pm$4.34 } & 1.05{ \small $\pm$0.09 } & 91.16{ \small $\pm$1.28 } & 0.42{ \small $\pm$0.02 } & 70.61{ \small $\pm$2.11 } & 1.25{ \small $\pm$0.09 } \\ 
    NO, cosine-like, ReLU & 87.48{ \small $\pm$1.81 } & 0.39{ \small $\pm$0.01 } & 71.19{ \small $\pm$1.14 } & 1.16{ \small $\pm$0.07 } & 89.54{ \small $\pm$0.18 } & 0.40{ \small $\pm$0.01 } & 68.86{ \small $\pm$4.35 } & 1.28{ \small $\pm$0.12 } \\ 
    NO, cosine-like, GELU & 87.89{ \small $\pm$0.24 } & 0.38{ \small $\pm$0.02 } & 71.36{ \small $\pm$3.22 } & 1.04{ \small $\pm$0.06 } & 89.84{ \small $\pm$1.61 } & 0.43{ \small $\pm$0.02 } & 73.76{ \small $\pm$2.33 } & 1.27{ \small $\pm$0.09 } \\ 
    NO, cosine-like, SwiGLUFFN & 89.79{ \small $\pm$1.38 } & 0.42{ \small $\pm$0.02 } & 69.13{ \small $\pm$4.66 } & 1.17{ \small $\pm$0.07 } & 89.70{ \small $\pm$2.29 } & 0.40{ \small $\pm$0.01 } & 71.21{ \small $\pm$4.02 } & 1.18{ \small $\pm$0.04 } \\ 
    NO, absolute linear, ReLU & 88.20{ \small $\pm$0.56 } & 0.46{ \small $\pm$0.03 } & 68.25{ \small $\pm$3.40 } & 1.02{ \small $\pm$0.05 } & 86.55{ \small $\pm$0.95 } & 0.43{ \small $\pm$0.04 } & 65.49{ \small $\pm$2.12 } & 1.13{ \small $\pm$0.11 } \\ 
    NO, absolute linear, GELU & 88.82{ \small $\pm$1.02 } & 0.48{ \small $\pm$0.02 } & 66.01{ \small $\pm$3.18 } & 1.06{ \small $\pm$0.07 } & 87.04{ \small $\pm$2.13 } & 0.43{ \small $\pm$0.02 } & 66.59{ \small $\pm$5.01 } & 1.12{ \small $\pm$0.11 } \\ 
    NO, absolute linear, SwiGLUFFN & 87.63{ \small $\pm$2.42 } & 0.44{ \small $\pm$0.03 } & 67.11{ \small $\pm$3.67 } & 1.11{ \small $\pm$0.07 } & 84.92{ \small $\pm$1.56 } & 0.43{ \small $\pm$0.01 } & 63.50{ \small $\pm$2.79 } & 1.12{ \small $\pm$0.10 } \\ 
 \hline 
    UB, relative linear, ReLU & 88.40{\small $\pm$1.32 } & 0.40{\small $\pm$0.04 } & 64.34{\small $\pm$3.66 } & 1.08{\small $\pm$0.13 } & 85.35{\small $\pm$2.05 } & 0.37{\small $\pm$0.02 } & 66.96{\small $\pm$3.12 } & 1.18{\small $\pm$0.08 } \\ 
    UB, relative linear, GELU & 89.06{\small $\pm$1.10 } & 0.43{\small $\pm$0.00 } & 69.70{\small $\pm$2.80 } & 1.26{\small $\pm$0.13 } & 86.94{\small $\pm$0.21 } & 0.37{\small $\pm$0.03 } & 68.79{\small $\pm$0.88 } & 1.22{\small $\pm$0.11 } \\ 
    UB, relative linear, SwiGLUFFN & 87.20{\small $\pm$4.98 } & 0.39{\small $\pm$0.02 } & 67.74{\small $\pm$7.17 } & 1.09{\small $\pm$0.15 } & 87.36{\small $\pm$1.32 } & 0.38{\small $\pm$0.04 } & 70.18{\small $\pm$1.84 } & 1.24{\small $\pm$0.05 } \\ 
    UB, cosine-like, ReLU & 87.92{\small $\pm$1.08 } & 0.39{\small $\pm$0.01 } & 68.85{\small $\pm$3.45 } & 1.17{\small $\pm$0.06 } & 84.54{\small $\pm$0.73 } & 0.37{\small $\pm$0.05 } & 68.02{\small $\pm$0.48 } & 1.25{\small $\pm$0.04 } \\ 
    UB, cosine-like, GELU & 89.63{\small $\pm$0.37 } & 0.40{\small $\pm$0.02 } & 73.79{\small $\pm$3.85 } & 1.13{\small $\pm$0.09 } & 85.86{\small $\pm$3.23 } & 0.39{\small $\pm$0.04 } & 67.61{\small $\pm$1.31 } & 1.35{\small $\pm$0.08 } \\ 
    UB, cosine-like, SwiGLUFFN & 87.91{\small $\pm$2.04 } & 0.39{\small $\pm$0.02 } & 68.99{\small $\pm$4.40 } & 1.19{\small $\pm$0.15 } & 87.95{\small $\pm$1.75 } & 0.39{\small $\pm$0.04 } & 67.54{\small $\pm$3.53 } & 1.23{\small $\pm$0.03 } \\ 
    UB, absolute linear, ReLU & 86.98{\small $\pm$1.99 } & 0.46{\small $\pm$0.02 } & 62.98{\small $\pm$0.75 } & 1.14{\small $\pm$0.10 } & 84.70{\small $\pm$4.27 } & 0.43{\small $\pm$0.02 } & 64.64{\small $\pm$1.86 } & 1.23{\small $\pm$0.08 } \\ 
    UB, absolute linear, GELU & 86.35{\small $\pm$0.47 } & 0.46{\small $\pm$0.01 } & 62.56{\small $\pm$5.20 } & 1.06{\small $\pm$0.07 } & 86.53{\small $\pm$2.16 } & 0.43{\small $\pm$0.02 } & 63.76{\small $\pm$1.08 } & 1.07{\small $\pm$0.02 } \\ 
    UB, absolute linear, SwiGLUFFN & 87.79{\small $\pm$0.74 } & 0.44{\small $\pm$0.03 } & 65.49{\small $\pm$3.42 } & 1.02{\small $\pm$0.03 } & 87.27{\small $\pm$1.57 } & 0.44{\small $\pm$0.04 } & 65.20{\small $\pm$4.51 } & 1.05{\small $\pm$0.05 } \\ 
  \bottomrule
  \end{tabular}

	\label{tab:ablation_study}
\end{table*}

\begin{table}[t]
	  \label{tab:table_02}
  \begin{tabular}{l|cc|cc}
    \toprule
    & \multicolumn{2}{c}{Latte-MV } & \multicolumn{2}{c}{TTHQ} \\
    Model & F1 Score & L1 Diff & F1 Score & L1 Dff \\
    \midrule
    NO Base cos & 86.0{\small $\pm$2.55} & 0.4{\small $\pm$0.02} & 46.4{\small $\pm$0.85} & 0.6{\small $\pm$0.06} \\ 
    NO EN cos & 89.8{\small $\pm$1.61} & 0.4{\small $\pm$0.02} & 73.8{\small $\pm$2.33} & 1.3{\small $\pm$0.09} \\ 
    NO Base absLin & 80.8{\small $\pm$1.56} & 0.5{\small $\pm$0.03} & 44.5{\small $\pm$0.28} & 0.5{\small $\pm$0.02} \\ 
    NO EN absLin & 87.0{\small $\pm$2.13} & 0.4{\small $\pm$0.02} & 66.6{\small $\pm$5.01} & 1.1{\small $\pm$0.11} \\ 
 \hline 
    UB Base cos & 88.3{\small $\pm$0.33} & 0.4{\small $\pm$0.08} & 36.1{\small $\pm$2.89} & 0.6{\small $\pm$0.07} \\ 
    UB EN cos & 85.9{\small $\pm$3.23} & 0.4{\small $\pm$0.04} & 67.6{\small $\pm$1.31} & 1.3{\small $\pm$0.08} \\ 
    UB Base absLin & 85.0{\small $\pm$0.14} & 0.5{\small $\pm$0.05} & 35.5{\small $\pm$6.09} & 0.6{\small $\pm$0.10} \\ 
    UB EN absLin & 86.5{\small $\pm$2.16} & 0.4{\small $\pm$0.02} & 63.8{\small $\pm$1.08} & 1.1{\small $\pm$0.02} \\ 
 \hline 
    NO EN cos 3d fps  & 89.8 $\pm${\small 1.61} & 0.4 $\pm${\small 0.02} & 73.8 $\pm${\small 2.33} & 1.3 $\pm${\small 0.09} \\ 
    NO EN cos 3d --  & 92.6 $\pm${\small 0.49} & 0.4 $\pm${\small 0.02} & 70.5 $\pm${\small 2.10} & 1.0 $\pm${\small 0.06} \\ 
    NO EN cos -- fps  & 92.3 $\pm${\small 0.27} & 0.4 $\pm${\small 0.02} & 55.5 $\pm${\small 3.75} & 0.9 $\pm${\small 0.02} \\ 
    NO EN cos -- --  & 94.0 $\pm${\small 0.09} & 0.4 $\pm${\small 0.00} & 61.0 $\pm${\small 3.72} & 0.9 $\pm${\small 0.22} \\ 
  \bottomrule
  \end{tabular}

	\caption{Performance comparison of the TCN baseline (Base) with EventNet (EN, mask tokens for NaN values + fps and viewpoint augmentation) using the original absolute linear (absLin) and our cosine-like (cos) time-to-event signal (top/middle section using keypoint set NO/UB). Bottom section shows the effect if we use or one of the two data augmentation strategies.}
	\label{tab:cnn_vs_transformer}
\end{table}

\subsection{Experiments Results and Ablation Study}
In this section, we systematically evaluate each proposed component and its various implementation options. The experiments are structured to validate our key design choices and demonstrate its performance. We start from the following default configuration: All models are exclusively trained on Latte-MV with a sequence length of $T=29$ using Adam as the optimizer and tested on Latte-MV's and TTHQ's test set. We further use fps and viewpoint augmentation, a maximum allowed frame distance of $\Delta t = 3$ and ReLUs. The input keypoints consist of the ball center, the six table keypoints, and 2 times the 9 upper body keypoints of both players.

\subsubsection{TCN-based Baseline vs. Our EventNet}
We compare the performance of the TCN baseline with our transformer-based EventNet using the absolute linear time-to-event encoding ("\emph{absLin}" for short) introduced by the baseline and our novel cosine-like signal ("\emph{cos}" for short) in the top two sections of Table~\ref{tab:cnn_vs_transformer}. The first section shows the results for the keypoint set without human poses (NO), while the second section with all keypoints of the upper bodies of both players (UB). We can see that our EventNet on Latte-MV test set outperforms the baseline on keypoint set NO by absolute $+6.2\%$ and $+3.8\%$ on "absLin" and "cos", respectively, while on keypoint set UB by $+1.5\%$ and $-2.4\%$, respectively. The avg. L1 distance slightly improves from $0.4$ to $0.5$ for the baseline to a constant $0.4$ with EventNet. On the challenging TTHQ test set we see an increase in absolute F1 score by EventNet over the baseline between $22.1\%$ and $31.5\%$. Furthermore, the transformer-architecture significantly accelerated training and inference by reducing the runtime between a factor of 0.32 and 0.48 on various GPU-platforms (e.g. with a single Nvidia RTX 3090, RTX 4090, A100, and Apple M4 Max). 

\subsubsection{Ablation Study}
Table \ref{tab:ablation_study} shows test results on Latte-MV and TTHQ using many variants of our EventNet architecture. In Table \ref{tab:ablation_study}, all models are exclusively trained on Latte-MV using fps and viewpoint augmentation, a sequence length of $T=29$, a maximum allowed frame distance of $\Delta t = 3$, and Adam as optimizer.

\paragraph{Temporal Event Encoding}
 If we look at the F1 test score on the Latte-MV dataset using the better performing keypoint set NO in Table \ref{tab:ablation_study} and average within each temporal event encoding type the three scores for the different FFN incarnations, we see that the relative linear target signal performs overall best with an avg. F1-Score of 89.00/ 89.91 over the cosine-like with 88.39/ 89.69 and absolute linear with 88.22 / 86.17 for masking / mask tokens for NaN values. However, its average mean L1 frame difference is minimally higher with a value of 0.42 / 0.41 versus 0.40 / 0.41 for cosine-like, but much better than absolute linear with
 0.46 / 0.43. This clearly shows the advantage of both relative target event signals. The effect is more pronounced on TTHQ, where both relative temporal target signal encodings outperform absolute linear by about $3.5\%$.

\paragraph{FFN Variants}
 If we analyze the F1 test scores on the Latte-MV dataset in Table \ref{tab:ablation_study} for our best configuration (NO, mask tokens for NaN values), and average the three values for the different temporal event encoding types, we observe that GELU performs best on Latte-MV. It achieves an average F1 test score of 88.62, surpassing ReLU and SwiGLUFFN marginally with scores of 88.56 and 88.59. On the challenging TTHQ dataset, the gap is a little bit larger with 69.68 compared to  69.14 and 68.44 for GELU and SwiGLUFFN. Given that TTHQ serves as a reliable proxy for out-of-training-domain performance (with variable frame rates and viewing angles), GELU emerges as the most suitable training option here.

\paragraph{Keypoint Set}
In table tennis, the ball isn't always easy to see or visible. Therefore, one idea was to use the players' upper-body joints as reference keypoints to help determine the timing of the ball-racket contact more reliably and, potentially, more accurately. 
We tested therefore whether the keypoint set UB outperforms the keypoints set NO. Table \ref{tab:ablation_study} gives here a clear answer. If the focus is on the best performing configuration using mask tokens for NaN values and averaging over all time-to-event signal and FFN variants, we see that NO outperforms UB by 2.31\% (88.59 vs. 86.28) on Latte-MV and by 2.12\% (69.09 vs. 66.97) on TTHQ. Similar results can be determined for the choice of mask NaN values. Note also that the average L1 difference is slightly lower for keypoint set NO than UB. Thus, body pose does not help in our specific event detection task. Probably, the reversal of the ball's direction provides the clearest signal for the example event detection task chosen.

\paragraph{FPS and Viewpoint Augmentation with Random Cameras}
The third and last section of Table \ref{tab:cnn_vs_transformer} shows the effect of our fps and viewpoint augmentations on inference performance. 
Contrary to expectations, fps augmentation does not consistently help. While it is clear that it should slightly degrade the performance on Latte-MV with consistent 30 fps videos, it only shows its benefit on TTHQ if combined with viewpoint augmentation. Here, the F1 score of 61.0\% without any augmentation increases to 70.5\% with viewpoint augmentations and further absolute 3.2 points to 73.8\% by adding also fps augmentations. Thus, viewpoint augmentations are highly effective to adapt to different viewpoints, while fps augmentation needs viewpoint augmentations to show its benefit.

\section{Conclusion}
We have presented a practical approach to event detection in table
tennis that combines compact 2D keypoint representations with transformer-based temporal modeling. In contrast to relying on raw video input directly, 2D keypoints are very efficient and generalize to diverse environmental settings very well. Therefore, EventNet is a modular solution for sports analysis in challenging video settings. 

Several central findings were discovered in this work. (1) A smartly designed small transformer-based architecture (using a keypoint and a time-to-event encoder) outperforms (speed- and performance-wise) a temporal convolutional network (TCN) substantially. (2) If the training data set is recorded at a fixed fps, the proposed fps data augmentation and a relative temporal time-to-event indicator signal (either relative linear or cosine-like) can transfer to variable fps rates of test videos. (3) If the training data set is recorded from a fixed camera view, the proposed viewpoint augmentation and a relative temporal time-to-event indicator signal can transfer to variable viewpoints of test videos. However, there are viewpoint cases where the event is not visible and therefore the overall temporal precision suffers (e.g., frontal views in table tennis). (4) The upper body pose of the table tennis player did not help in the case of ball-racket contact estimation, except in  cases where the ball is not visible such as for frontal view recordings. In all other cases, the upper body pose degraded the temporal precision in time as the ball - if visible - is the best indicator for identifying the point of time when the ball changes its direction. Despite that we focus on racket-ball contact in this paper, our approach is transferable to all kinds of motion event detection.

\begin{acks}
The authors gratefully acknowledge the resources on the LiCCA HPC cluster of the University of Augsburg, co-funded by the Deutsche Forschungsgemeinschaft (DFG, German Research Foundation) – Project-ID 499211671. This work was partly supported by JSPS Invitational Fellowship for Research in Japan (Long-Term L26503) and by JST ASPIRE JPMJAP2303.
\end{acks}

\bibliographystyle{ACM-Reference-Format}
\bibliography{mmsports2026}

\end{document}